\documentclass{article}

\usepackage{microtype}
\usepackage{graphicx}
\usepackage{subfigure}
\usepackage{booktabs} 
\usepackage{amsmath}
\usepackage{adjustbox}
\usepackage{multirow}
\usepackage{siunitx}
\usepackage{listings}
\usepackage{xcolor} 
\usepackage{enumitem}

\usepackage{hyperref}

\usepackage[accepted]{icml2025}

\usepackage{amsmath}
\usepackage{amssymb}
\usepackage{mathtools}
\usepackage{amsthm}

\usepackage[capitalize,noabbrev]{cleveref}

\theoremstyle{plain}

\theoremstyle{definition}

\theoremstyle{remark}

\newtheoremstyle{boldstyle} 
  {10pt} 
  {10pt} 
  {\itshape} 
  {} 
  {\bfseries} 
  {:} 
  {.5em} 
  {} 

\theoremstyle{boldstyle}
\newtheorem{property}{Property}

\usepackage[textsize=tiny]{todonotes}

\begin{document}

\newcommand{\name}{{CommVQ}}

\twocolumn[
\icmltitle{{\name}: Commutative Vector Quantization for KV Cache Compression}



\icmlsetsymbol{equal}{*}

\begin{icmlauthorlist}
\icmlauthor{Junyan Li}{umass}
\icmlauthor{Yang Zhang}{mit}
\icmlauthor{Muhammad Yusuf Hassan}{umass}
\icmlauthor{Talha Chafekar}{umass}
\icmlauthor{Tianle Cai}{princeton}
\icmlauthor{Zhile Ren}{apple}
\icmlauthor{Pengsheng Guo}{apple}
\icmlauthor{Binazir Karimzadeh}{apple}
\icmlauthor{Colorado J Reed}{apple}
\icmlauthor{Chong Wang}{apple}
\icmlauthor{Chuang Gan}{umass}
\end{icmlauthorlist}

\icmlaffiliation{umass}{University of Massachusetts Amherst}
\icmlaffiliation{mit}{Massachusetts Institute of Technology}
\icmlaffiliation{princeton}{Princeton University}
\icmlaffiliation{apple}{Apple Inc}

\icmlcorrespondingauthor{Junyan Li}{junyanli@umass.edu}

\icmlkeywords{Machine Learning, ICML}

\vskip 0.3in
]



\printAffiliationsAndNotice{}  

\begin{abstract}

Large Language Models (LLMs) are increasingly used in applications requiring long context lengths, but the key-value (KV) cache often becomes a memory bottleneck on GPUs as context grows. To address this, we propose \textbf{Comm}utative \textbf{V}ector \textbf{Q}uantization ({\name}) to significantly reduce memory usage for long-context LLM inference. We first introduce additive quantization with a lightweight encoder and codebook to compress the KV cache, which can be decoded via simple matrix multiplication. To further reduce computational costs during decoding, we design the codebook to be commutative with Rotary Position Embedding (RoPE) and train it using an Expectation-Maximization (EM) algorithm. This enables efficient integration of decoding into the self-attention mechanism. Our approach achieves high accuracy with additive quantization and low overhead via the RoPE-commutative codebook. Experiments on long-context benchmarks and GSM8K show that our method reduces FP16 KV cache size by 87.5\% with 2-bit quantization, while outperforming state-of-the-art KV cache quantization methods. Notably, it enables 1-bit KV cache quantization with minimal accuracy loss, allowing a LLaMA-3.1 8B model to run with a 128K context length on a single RTX 4090 GPU. The source code is available at: \url{https://github.com/UMass-Embodied-AGI/CommVQ}.

\end{abstract}


\section{Introduction}
\label{sec:intro}
We are witnessing a growing trend in increasing the context length of large language models (LLMs). For instance, the latest LLaMA 3.1 models \citep{dubey2024llama} support up to a 128K context length, and recent research \citep{ding2024longrope,jin2024llm} has managed to extend this even further, with some models achieving context lengths of over 1 million. Supporting longer contexts enables LLMs to process richer inputs and generate more tokens, improving their capacity for more complex tasks and reasoning \citep{wei2022chain}.

However, this increased context length presents a significant challenge on GPU memory usage. The causal attention mechanism used in LLMs relies on a Key-Value (KV) cache to speed up inference. This cache stores the keys and values of all previous tokens, eliminating the need to recompute them when generating the next token. As context lengths increase, the size of the KV cache grows proportionally, eventually becoming the primary bottleneck for memory usage — often far exceeding the memory required for the model itself. For instance, a LLaMA 3.1 8B model requires 16 GB of memory to store the model weight in FP16 precision. If the context length is set to its maximum of 128K with a batch size of 2, the KV cache alone would require 88 GB of memory. This makes it impossible to run inference on a single GPU without KV cache offloading, even for H100-80GB.

Efforts to reduce KV cache size are ongoing \citep{shi2024keep,yuan2024kv}, with KV cache quantization \citep{liu2024kivi,hooper2024kvquant} being a key approach. Quantization lowers the memory footprint by reducing the bit width used to represent each FP16 scalar in the KV cache. For instance, INT4 quantization cuts memory usage by a factor of four compared to FP16. However, this comes at a cost: aggressive quantization, such as 2-bit or even 1-bit quantization, results in significant information loss, severely degrading model performance.

\newcommand{\footstar}[1]{%
  \renewcommand{\thefootnote}{}
  \footnotetext{$^*$#1}%
  \addtocounter{footnote}{-1}
}
\footstar{All experiments were conducted by Junyan Li at UMass Amherst.}

To address these challenges, we propose \textbf{Comm}utative \textbf{V}ector \textbf{Q}uantization ({\name}), a novel method for efficient and accurate KV cache quantization tailored to long-context LLMs. Unlike existing quantization techniques that treat each scalar in the KV cache independently, {\name} performs quantization at the vector level. Specifically, we treat the key/value vector for each token as a single unit rather than processing scalars individually. To achieve this, we leverage additive quantization \citep{babenko2014additive}, a variant of vector quantization, to encode each vector into a low-bitwidth representation utilizing a learned codebook, minimizing quantization error while significantly reducing memory usage.

More importantly, to integrate additive quantization into the self-attention mechanism in a computationally efficient manner, the codebook is innovatively designed to be commutative with the Rotary Position Embedding (RoPE) matrix \citep{su2024roformer}. This allows a drastic reduction in the computational overhead of KV decoding, where intermediate results can be pre-computed against each code in the codebook, and are then efficiently reused in computing the lengthy key-query products.

By combining these innovations, {\name} achieves a superior trade-off between memory savings and accuracy. Extensive evaluation on two long-context benchmarks, LongBench~\citep{bai2023longbench} and InfiniteBench~\citep{zhang2024bench}, as well as GSM8K~\citep{cobbe2021training}, a benchmark designed for complex reasoning, shows that compared to state-of-the-art KV cache quantization baselines, we achieve nearly lossless KV cache compression with 2-bit quantization, outperforming other methods. Furthermore, we achieve 1-bit quantization with significantly better accuracy than existing baselines, demonstrating the effectiveness of our method in pushing the compression limits of KV cache. We summarize our contributions as follows:
\begin{itemize}
    \item Unlike prior work that quantizes each scalar individually in the KV cache, we quantize each \textbf{vector as a whole} in the KV cache into a low bit-width representation using a learned codebook.
    
    \item By leveraging the \textbf{commutative property} of the RoPE matrix and the characteristics of self-attention, we refine our codebook to be \textbf{RoPE-commutative}. This refinement enables us to reformulate the self-attention computation to incorporate the decoding process more efficiently. Additionally, we provide a Triton implementation of our method to demonstrate real memory savings.
    
    \item Extensive experiments demonstrate the superiority of our method, particularly in \textbf{ultra-low-bitwidth} KV cache quantization scenarios (\emph{e.g.}, 1-bit quantization). This opens up the possibility of serving long-context LLMs under limited GPU memory constraints.
\end{itemize}
\section{Related Works}
\label{sec:related}
\noindent\textbf{KV Cache Compression.} Several prior works have addressed KV cache compression, which generally fall into two categories: token eviction and quantization. Methods focused on token eviction \citep{xiao2024duoattention, liu2024scissorhands, zhang2023h2o} aim to reduce KV cache size by evicting less important tokens, storing only the keys and values for the most relevant tokens. These approaches are orthogonal to our method and could potentially be combined with it to achieve even higher compression rates.

Another important approach to KV cache compression is quantization \citep{liu2024kivi, hooper2024kvquant, zhang2024kv, kumar2024residual}, which reduces the bit width of the KV cache, thereby lowering its overall storage requirements. While prior works have successfully demonstrated 4-bit and 2-bit quantization for KV caches, few have explored the feasibility of achieving 1-bit quantization for long-context LLMs. Our method addresses this gap by introducing a novel vector based KV cache quantization technique that enables 1-bit quantization with minimal accuracy loss. Furthermore, we offer a new perspective on how to reformulate self-attention mechanisms and our quantization method for a much more efficient integration.

\noindent\textbf{Vector Quantization.}
Vector quantization \citep{gray1984vector} is a widely studied technique in signal processing and machine learning that represents high-dimensional data using a smaller set of representative vectors, known as codebooks. Variants of VQ, such as product quantization (PQ) \citep{jegou2010product} and additive quantization (AQ) \citep{babenko2014additive}, have been introduced to improve its efficiency and capacity.

In machine learning, vector quantization has been successfully applied to areas such as generative modeling \citep{van2017neural, esser2021taming}. However, its potential extension in the context of KV cache compression, remains largely unexplored. A recent study, VQLLM \citep{kumar2024residual}, introduced residual vector quantization (RVQ) for compressing the KV cache. However, their achieved compression rate is relatively modest, and their basic decode-then-self-attention process introduces significant computational overhead, limiting its practicality for serving long-context LLMs. There remains a lack of in-depth research on how to optimize vector quantization and better integrate it with self-attention for both more efficient and effective KV cache compression.

\section{Preliminaries}
\label{sec:preliminary}

\subsection{Self-Attention and KV Cache}

Self-attention is the fundamental building block of LLMs. It takes query (Q), key (K), and value (V) matrices as inputs and produces an output (O) with the same shape as Q. The causal attention mask used in self-attention allows us to cache the key and value matrices, significantly speeding up token generation. During LLM inference, the process is divided into two stages:

\noindent\textbf{Prefilling Stage.} At this stage, given the input prompt, the attention output is computed while simultaneously generating the KV cache. Given the hidden states of the prompt, $X\in\mathbb{R}^{N\times d}$, where N is the number of tokens, and d is the hidden size, the Q, K, and V matrices, as well as the self-attention output, are computed as follows:
\begin{equation}
\begin{aligned}
    Q = XW_Q, \quad K = XW_K, \quad V = XW_V \\
    \text{Self-Attn} = \text{Softmax}\left(\frac{QK^T}{\sqrt{d}}\right)V  \label{eqn:self-attn}
\end{aligned}
\end{equation}
Here, $W_Q$, $W_K$, and $W_V \in \mathbb{R}^{d\times d}$ are the projection matrices for the query, key, and value, respectively. The computed K and V matrices are then cached for use in the subsequent decoding stage.

\noindent\textbf{Decoding Stage.} During this stage, the KV cache is reused for self-attention computations. Given the current input hidden state $x\in\mathbb{R}^{1\times d}$, the KV cache is updated as follows:
\begin{equation}
    K \leftarrow \text{Concat}(K, xW_K),V\leftarrow \text{Concat}(V, xW_V)
\end{equation}
The updated KV cache is then used to compute the self-attention output, following the same equation as in Eqn.~\ref{eqn:self-attn}:
\begin{equation}
\begin{aligned}
    Q &= xW_Q \\
    \text{Self-Attn} &= \text{Softmax}\left(\frac{QK^T}{\sqrt{d}}\right)V 
\end{aligned}
\end{equation}

The prefilling stage occurs once to process the input tokens and generate the first output token. This is followed by multiple iterations of the decoding stage, which produces all subsequent output tokens.

\subsection{Rotary Position Embedding}
\label{sec:rope}

Positional encoding is added to the query (Q) and key (K) matrices to encode token position information during self-attention computation. Recent open-source LLMs, such as LLaMA \citep{dubey2024llama}, Mistral \citep{jiang2023mistral}, and Qwen \citep{yang2024qwen2}, commonly utilize Rotary Position Embedding (RoPE) \citep{su2024roformer} for this purpose:
\begin{equation}
    q_m \leftarrow q_mR_m, \quad k_m \leftarrow k_mR_m
\end{equation}

where $q_m, k_m \in \mathbb{R}^{1\times d}$ represent the query and key vector for the $m^{\text{th}}$ token, respectively, while $R_m \in \mathbb{R}^{d\times d}$ denotes the RoPE matrix applied to the $m^{\text{th}}$ token. RoPE matrix is a sparse matrix with nonzero values only in its $2 \times 2$ diagonal blocks. Therefore, the application of RoPE can be reformulated by first dividing the \( k_m \) vector (and similarly \( q_m \); here, we use \( k_m \) as an example) into multiple 2-dimensional sub-vectors:
\begin{equation}
\begin{aligned}
    k_m &= [k_{1x},k_{1y},...,k_{(d/2)x},k_{(d/2)y}] \\
        &= [k_m^1,...,k_m^{d/2}]
    \label{eqn:divide_k}
\end{aligned}
\end{equation}
where each 2-dimensional sub-vector $k_m^i = (k_{ix}, k_{iy})$ consists of two scalars within the $k_m$ vector. The corresponding $2\times 2$ diagonal sub-matrix of $R_m$, denoted as $R_m^i \in \mathbb{R}^{2\times2}$, is then applied to each sub-vector $k_m^i$
\begin{equation}
\begin{aligned}
R_{m}^i &= \left(\begin{array}{cc}
\cos m \theta_i & -\sin m \theta_i \\
\sin m \theta_i & \cos m \theta_i \\
\end{array}\right) \label{eqn:rope_ri}\\
k_m^i &\leftarrow k_m^iR_{m}^i
\end{aligned}
\end{equation}

where $\theta_i=10000^{-2(i-1)/d}$. The $2 \times 2$ diagonal sub-matrix $R_m^i$ is, in fact, a rotation matrix, which satisfies the following property:

\begin{property}[Commutativity]\label{prop:commutativity}
Let $C\in\mathbb{R}^{2\times2}$ defined as
\begin{equation}
C = 
\begin{pmatrix}
    x & y \\
    -y & x \\
\end{pmatrix}
\label{eqn:C_struct}
\end{equation}
The following commutativity holds:
\begin{equation}
    R_m^iC=CR_m^i
\end{equation}
\end{property}

This property shows that \( C \) and \( R_m^i \) are commutative under matrix multiplication. We leverage this key property to optimize our method in Sec.~\ref{sec:rope_aware_key}.

\section{Method}
\label{sec:method}

The KV cache is a dominant factor in long-context LLM inference scenarios, as storing and loading the large KV cache becomes a significant memory and latency bottleneck. Therefore, compressing the KV cache, even at the cost of additional encoding and decoding processes, is advantageous. Motivated by vector quantization, a classical quantization technique from signal processing, and particularly inspired by its recent variant, additive quantization~\citep{babenko2014additive}, we adopt a similar approach to quantize each vector in the KV cache into a compressed representation using a learned encoder and codebook, as described in Sec.~\ref{sec:add_quant}.

To efficiently integrate vector quantization into self-attention while minimizing the computational overhead introduced by the additional decoding process, we innovatively reformulate the self-attention computation by designing a RoPE-commutative codebook and reordering the matrix multiplications involved in self-attention. This refinement enables a large portion of computation reuse and significantly reduces the computational cost of our method, as detailed in Sec.~\ref{sec:rope_aware_key}.

\subsection{Learning Additive Quantization for KV Cache}
\label{sec:add_quant}

\begin{figure}[ht]
    \centering
    \includegraphics[width=0.9\columnwidth]{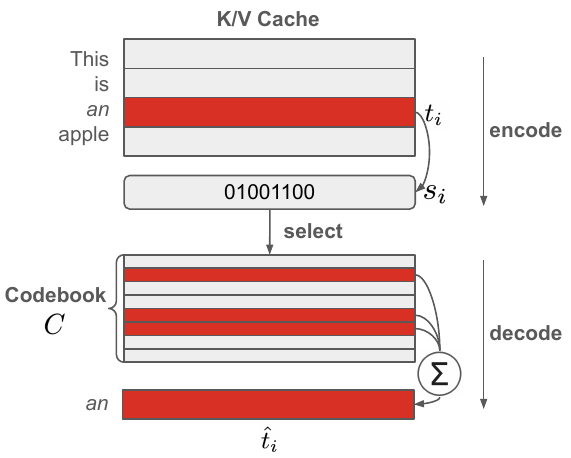}
    \caption{\textbf{An illustration of vector quantization for KV cache compression.} Each vector in the KV cache, corresponding to a token, is first encoded into a low-bitwidth representation, significantly reducing storage size. This process is applied to all vectors in the KV cache individually. When needed, the compressed representations are loaded and decoded back to the original KV cache using a codebook.}
    \label{fig:method}
\end{figure}

Additive quantization~\citep{babenko2014additive} aims to represent a given \( d \)-dimensional vector as the element-wise summation of several vectors from a learned codebook $C$. For simplicity, we adopt a per-token quantization scheme for both the key and value matrices, meaning that the \( d \)-dimensional key and value vectors are quantized individually for each token.

Inspired by additive quantization, our method incorporates a learned encoder \( E \) to \textbf{encode} the given \( d \)-dimensional vector into a binary sequence, consisting of 0s and 1s, of length \( N_c \), and a codebook \( C \in \mathbb{R}^{N_c \times d} \) to \textbf{decode} the original vector back using a simple matrix multiplication, as illustrated in Figure~\ref{fig:method}.

\noindent\textbf{Encoding.} Given a key or value vector for the \( i^{\text{th}} \) token, \( t_i \in \mathbb{R}^d \), we use an encoder \( E \) to encode this vector into a quantized vector \( s_i \in \{0,1\}^{N_c} \), namely $s_i = E(t_i)$. In our implementation, the encoder \( E \) consists of a simple linear layer followed by an activation function, and then another linear layer for output. Gumbel-softmax~\citep{jang2016categorical} is used to make the encoder end-to-end differentiable. The encoding step takes place after the generation of the KV cache, and the quantized vector \( s_i \) for each token is concatenated and stored together to form the quantized KV cache \( S \).

\noindent\textbf{Decoding.} When loading the KV cache, the quantized KV cache \( S \) is retrieved. For each token's \( s_i \), we perform a simple matrix multiplication with the codebook $C$ to reconstruct the decoded tensor \( \hat t_i \):
\begin{equation}
    \hat t_i = s_i C
    \label{eqn:vanilla_decode}
\end{equation}
The decoded key and value can then participate in the subsequent computations in the self-attention mechanism.

The encoder $E$ and codebook $C$ are optimized via gradient descent to minimize the MSE loss between the original tensor $t_i$ and its decoded counterpart $\hat t_i$.

\noindent\textbf{KV Cache Reduction Analysis.} The size of the original FP16 KV cache for each layer can be calculated as \( B \times N \times d \times 2 \times 16 \) bits, where B is the batch size, N is the number of tokens and d is the hidden size. After quantization, the size of the quantized KV cache for each layer becomes \( B \times N \times N_c \times 2 \times 1 \) bits. Therefore, the reduction rate \( RR \) can be expressed as \( 1 - \frac{N_c}{16d} \). Subsequently, the average bit width used to represent each scalar in the KV cache is:
\begin{equation}
    \textbf{Avg. bit} = 16 - 16 \times RR = \frac{N_c}{d}
\end{equation}
For instance, the hidden size $d$ for the key and value for a LLaMA-3.1-8B-Instruct model is $1024$. If we set \( N_c \) to $1024$, the reduction rate will be \( \frac{15}{16} \), which is equivalent to 1-bit quantization. We will use \textbf{Avg. bit} as the metric to measure the compression rate in our experiment section. A lower \textbf{Avg. bit} indicates a higher compression rate and more reduced GPU memory usage compared to FP16.

\noindent\textbf{Computation Complexity.} The computational overhead arises from both encoding and decoding the KV cache. Since the KV cache only needs to be encoded once, the primary source of computational overhead comes from decoding the full KV cache each time it is used during the generation of the next token.

The computation for self-attention with KV cache decoding during inference step $t$ can be summarized as:
\begin{align}
    q &\leftarrow qR_t \\
    K &\leftarrow \begin{bmatrix}
        s_0C_KR_0 \\
        s_1C_KR_1 \\
        ... \\
        s_{N-1}C_KR_{N-1}
    \end{bmatrix} \\
    V &\leftarrow S_VC_V \label{eqn:decode_v} \\
    \text{Self-Attn} &= \text{Softmax}\Big(\frac{qK^T}{\sqrt{d}}\Big)V
    \label{eqn:v1_formula}
\end{align}

where \( q \in \mathbb{R}^{1 \times d} \) is the current query, $N$ is the number of tokens generated so far, \( S_V \in \{0,1\}^{N \times N_c} \) are the quantized value, $s_i \in \{0,1\}^{N_c}$ is the quantized vector for token-$i^{th}$ key, \( C_K, C_V \in \mathbb{R}^{N_c \times d} \) are the codebook for the key and value, respectively, and $R_i$ denote the RoPE matrix for $i^{th}$ token. The computational complexity can be expressed as:
\begin{equation}
    \label{eqn:v1_complexity}
    O\left((2d+1)N+2dN_cN\right)
\end{equation}
where \( d \) is the hidden dimension, \( N \) is the number of tokens, and \( N_c \) is the number of rows in the codebook. The first term in Eqn.~\ref{eqn:v1_complexity} corresponds to the computation for Eqn.~\ref{eqn:v1_formula}, which represents vanilla self-attention. The second term, \( 2dN_cN \), accounts for the KV cache decoding process for the key and value, where each contributes \( dN_cN \).  

Comparing the first and second terms in Eqn.~\ref{eqn:v1_complexity}, we see that the computational overhead from the additional vector quantization decoding process is \( N_c \) times higher than that of vanilla self-attention. This increase arises from the need to first sum a large number of rows in the codebook to reconstruct the decoded vector, after which the decoded vector can participate in the self-attention computation. The overhead is particularly significant because $N_c$ is typically large (\textit{e.g.}, $1024$ for 1-bit quantization in LLaMA-3.1 8B model).

In the next section, we introduce our novel redesign, which leverages commutative codebooks to more efficiently integrate vector quantization with self-attention. This method will significantly reduce the overall computational overhead.

\subsection{Commutative Codebook for Efficiency}
\label{sec:rope_aware_key}

To simplify notation, we focus on the calculation within the softmax and ignore the $\sqrt{d}$ constant:  
\( \alpha = qK^T \), where the output \( \alpha \) is an \( N \)-dimensional vector. Each scalar entry \( \alpha_i \) in $\alpha$ is computed as:  
\begin{equation}
    \alpha_i = qR_t(s_iC_KR_i)^T = (qR_t)R_i^T C_K^T s_i^T  \label{eqn:attention}
\end{equation}

where \( R_t \) denotes the RoPE matrix for the query, \( R_i \) denotes the RoPE matrix for the \( i^{th} \) key, and \( s_i \) is the quantized vector for the \( i^{th} \) key. Notice that \( qR_t \) and \( C_K \) remain unchanged as \( i \) varies. If \( R_i \) were independent of \( i \) as well, then the computation of \( (qR_t)R_i^T C_K^T \) would only need to be performed once across different \( i \), significantly reducing the computational cost. Unfortunately, since \( R_i \) varies with \( i \), such a reduction is not possible.

However, if we could design the codebook $C_K$ to be commutative with $R_i$, then the right-hand-side of Eqn.~\ref{eqn:attention} could be rewritten as \( (qR_t) C_K^T R_i^T s_i^T\), which makes the bulk of computation, \( (qR_t)C_K^T \), independent of $i$ and thus reusable, leading to substantial computational savings.

Next, we discuss how to define our new commutative codebook $C_K$ as well as how to learn it. 

\noindent\textbf{Designing a Commutative Codebook.} As explained in Sec.~\ref{sec:rope}, since the RoPE matrix is block-diagonal, we can break the problem into 2-dimensional subspaces and design the commutative codebook within each block. Formally, as defined in Eqns.~\ref{eqn:divide_k} and \ref{eqn:rope_ri},  $k_i^{j}$ and $R_i^{j}$ represent the $j$-th 2-dimensional sub-vector/sub-matrix of $k_i$ and $R_i$, respectively. Extending \textbf{Property~\ref{prop:commutativity}}, we can obtain a design for a sub-space codebook.

Specifically, let \(\mathcal{C}_K^j\) be a set of codebooks for subspace $j$ of the key vector, \emph{i.e.}, 
\begin{equation}
\mathcal{C}_K^j = \{C_K^{j0}, C_K^{j1}, \ldots, C_K^{j(N_{c'}-1)}\}
\end{equation}
where $N_{c'}$ is the number of quantization levels, and each $C_K^{jl}$ is a $2\times 2$ matrix that satisfies the form defined in Eqn.~\ref{eqn:C_struct}, and thus has the commutative property
$R_i^j C_K^{jl} = C_K^{jl} R_i^j$.

The quantized vector for $k_i^j$, denoted as $s_i^j$, is a 2-dimensional vector taking the values from $\{0, \ldots, N_{c'}-1\}$. The decoded key is represented as
\begin{equation}
    \hat{k}_i^j = \sum_{l=0}^{N_{c'}-1} [s_i^j = l] C_K^{jl}
    \label{eqn:commute_decode}
\end{equation}
where \( [s_i^j = l] \) is a 2-dimensional boolean indicator vector, with each dimension equal to 1 if the corresponding dimension of \( s_i^j \) is equal to \( l \), and 0 otherwise.

Notice that the decoding scheme in Eqn.~\ref{eqn:commute_decode} is more complicated than previously discussed (Eqn.~\ref{eqn:vanilla_decode}). This is because previously we used the same codebook for all the dimensions, but now the codebook for different dimensions is different. Please refer to Appendix~\ref{sec:more_detail} for more explanations.

With the new decoding scheme, the benefit of commutativity remains. To see this, notice that Eqn.~\ref{eqn:attention} should now be rewritten as
\begin{equation}
\begin{aligned}
    \alpha_i &= \sum_j (q^jR^j_t)(\hat{k}_i^jR^j_i)^T \\
    &= \sum_j (q^jR^j_t)\big(\sum_{l}[s_i^j = l] C_K^{jl}R^j_i\big)^T \\
    &= \sum_{j,l} (q^jR^j_t)R^{jT}_i C_K^{jlT}[s_i^j = l]^T \\
    &= \sum_{j,l} (q^jR^j_t) C_K^{jlT}R^{jT}_i [s_i^j = l]^T
    \label{eqn:new_attention}
\end{aligned}
\end{equation}
The first equality decompose the inner products into those of the sub-vectors; the last equality applies the commutativity property, making $(q^jR^j_t) C_K^{jlT}$ reusable across $i$.

\noindent\textbf{Learning the Codebook.} The codebook is learned by minimizing the reconstruction error:
\begin{equation}
    \min_{\cup_{i} (\mathcal{C}_K^j, s_i^j)} \sum_{i} \Vert \hat{k}_i^j - k_i^j\Vert^2, \text{s.t. Eqn. \ref{eqn:commute_decode}}
\end{equation}
This is a canonical clustering objective and can be efficiently solved via an EM-like algorithm, where the E-step minimizes over $s_i^j$ holding $\mathcal{C}_K^j$ fixed, and the M-step minimizes over $\mathcal{C}_K^j$ holding $s_i^j$ fixed, as described in \textbf{Algorithm~\ref{alg:em_algo}}. More details can be found in Appendix~\ref{sec:em_algo_impl_details}, including the actual update formula and the techniques used to stabilize the optimization process.

\begin{algorithm}[ht]
    \begin{algorithmic}[1]
    \STATE \textbf{Input:} A calibration set $K\in\mathbb{R}^{N\times 2}$ \\
    \STATE \textbf{Parameters:} codebook $\mathcal{C}_K^j$
    \STATE \textbf{Goal:} Optimize $\mathcal{C}_K^j$ to minimize the clustering error over calibration set $K$.
    \WHILE{$\mathcal{C}_K^j$ converges}
        \STATE \textbf{E Step:} Fix $\mathcal{C}_K^j$, update the assignment $S$ such that each $k\in K$ is assigned to its nearest clustering center.
        \STATE \textbf{M Step:} Fix $S$, update $\mathcal{C}_K^j$ such that the MSE loss between each $k$ and its assigned clustering center is minimized. 
    \ENDWHILE
    \end{algorithmic}
    \caption{\small EM Algorithm for Learning RoPE Commutative Codebook.}
    \label{alg:em_algo}
\end{algorithm}

Notice that $s_i^j$ requires $2\log_2(N_{c'})$ bits, which limits its ability to achieve a high compression rate relative to $k_i^j$. To improve compression, we group consecutive \( g \) sub-vectors into one group and share the quantized value within the group, \emph{i.e.}, $s_i^0 = s_i^1 = \dots = s_i^{g-1}$. This allows the entire vector to be represented using significantly fewer bits.

To further improve the quantization accuracy, we iteratively apply the clustering algorithm on the quantization error tensors with new codebook for each time, repeating the process until the error is sufficiently minimized. Specifically, we run the clustering algorithm \( R \) times, resulting in $R$ codebooks $\mathcal{C}_K^{j}$ for subspace $j$. $R$ is a hyperparameter that balances quantization accuracy and compression rate. This iterative refinement is conceptually similar to the residual approach in residual vector quantization \citep{barnes1996advances}, where subsequent iterations focus on reducing the remaining error. As a result, a total of \( R \) instances of \( s \) are required to quantize $2g$ FP16 scalars, and the total number of bits needed to quantize the full $d$-dimensional vector is $2R\log_2(N_{c'})\frac{d}{2g}$. Accordingly, the average quantization bit can be computed as:
\begin{equation}
    \textbf{Avg. bit} = \frac{R\log_2(N_{c'})}{g}
\end{equation}
For example, to achieve 1-bit quantization, we set $N_{c'} = 64$, $R=11$ and $g=64$. We provide an ablation study on how to choose $N_{c'}$, $R$ and $g$ in Appendix~\ref{sec:ablate_codebook}.

\noindent\textbf{Reduced Computational Complexity.} For value quantization, we retain our original method but reorder the matrix multiplication process. Specifically, we first multiply the softmaxed attention weights by \( S_V \), followed by multiplying the result by \( C_V \), as illustrated below:
\begin{equation}
    \text{Self-Attn} = \text{Softmax}(A)S_VC_V 
    \label{eqn:optimized_v}
\end{equation}
This simple reordering of matrix multiplication reduces the computational complexity from \( O(dN_cN + dN) \) in Eqns.~\ref{eqn:decode_v},\ref{eqn:v1_formula} to \( O(N_cN + dN_c) \) in Eqn.~\ref{eqn:optimized_v}, which is nearly \( d \) times lower, assuming \( d \) and \( N_c \) are of a similar scale.

By integrating these adjustments into the original self-attention mechanism, the optimized computation complexity is now:
\begin{equation}
    O((Rd + N_{c} + 1)N + d(N_{c} + RN_{c'}))
    \label{eqn:final_comp}
\end{equation}
This optimization effectively reduces the computational cost compared to the unoptimized decode-then-self-attention calculation (Eqn.~\ref{eqn:v1_complexity}).
 Previously, the complexity was $N_c$ times higher than that of the original self-attention; now, it is approximately $\frac{R+1}{2}$ times higher. Since $R$ is a relatively small hyperparameter (\textit{e.g.}, $R=11$ for 1-bit quantization), the overhead remains minimal.

\section{Experiments}
\label{sec:exp}
\subsection{Settings}

\noindent\textbf{Models.} We evaluate {\name} using the latest LLaMA-3.1-8B-Instruct model \citep{dubey2024llama}, which supports a context length of up to 128K tokens. A subset of the FineWeb-Edu dataset \citep{lozhkov2024fineweb-edu} is used to learn the encoder and codebooks. We present evaluation results for two quantization levels: 2-bit and 1-bit quantization (see Appendix~\ref{sec:ablate_codebook} for codebook configuration). To demonstrate the generalizability of our method, we also conduct additional experiments on the LLaMA-2-8B \citep{touvron2023llama} and Mistral-8B \citep{jiang2023mistral} models.

\noindent\textbf{Baselines.} We compare our method to three recent KV cache quantization techniques: KIVI~\citep{liu2024kivi}, KVQuant~\citep{hooper2024kvquant}, and VQLLM~\citep{kumar2024residual}. KIVI employs asymmetric quantization, KVQuant uses non-uniform quantization, and VQLLM applies residual vector quantization. For fairness, we reproduced their results using their official open-source implementations on the same models. We denote quantization versions as \textless method\textgreater-\textless n\textgreater, where \textless n\textgreater~represents bits per scalar in KV cache. For VQLLM, we set $C=256, K=8$ for 2-bit and $C=256, K=4$ for 1-bit quantization.

\begin{table*}[t]
\centering
\sisetup{
    table-format=2.2,
}
\begin{adjustbox}{width=1.99\columnwidth,center}
\begin{tabular}{lcSSSSSSSSS}
\toprule
\multicolumn{1}{l}{\textbf{Method}} & \textbf{Avg. bit} ($\downarrow$) & {\textbf{Qasper}} & {\textbf{QMSum}} & {\textbf{MultiNews}} & {\textbf{TREC}} & {\textbf{TriviaQA}} & {\textbf{SAMSum}} & {\textbf{LCC}} & {\textbf{RepoBench-P}} & {\textbf{Average} ($\uparrow$)} \\ \midrule
FP16 Baseline & 16 & 25.19 & 23.31 & 26.82 & 72.50 & 91.65 & 43.49 & 52.47 & 49.01 & 48.05 \\ \midrule
KIVI-2 & 3.00 & 22.71 & 24.33 & 27.29 & 72.50 & 92.06 & 43.26 & 51.32 & 47.53 & 47.62 \\
KVQuant-2 & 2.33 & 41.86 & 22.37 & 25.76 & 69.00 & 89.00 & 42.09 & 36.22 & 36.51 & 45.35 \\
VQLLM-2 & 2.00 & 32.39 & 25.20 & 26.22 & 69.95 & 92.01 & 41.03 & 40.58 & 36.19 & 45.45 \\
\textbf{{\name}-2} & \textbf{2.00} & 24.67 & 24.36 & 26.48 & 72.50 & 91.92 & 43.98 & 53.02 & 46.92 & \textbf{47.98} \\ \midrule
KIVI-1 & 2.00 & 4.99 & 9.57 & 9.20 & 38.75 & 25.07 & 11.93 & 17.67 & 16.40 & 16.70 \\
KVQuant-1 & 1.33 & 1.01 & 8.71 & 6.06 & 1.00 & 1.50 & 6.64 & 11.01 & 11.09 & 5.88 \\
VQLLM-1 & \textbf{1.00} & 11.92 & 17.91 & 13.12 & 47.98 & 63.34 & 23.72 & 18.92 & 22.44 & 27.42 \\
\textbf{{\name}-1} & 1.03 & 18.86 & 23.02 & 24.34 & 69.00 & 91.61 & 41.83 & 48.78 & 42.08 & \textbf{44.94} \\ 
\bottomrule
\end{tabular}
\end{adjustbox}
\caption{LongBench evaluation for the LLaMA-3.1-8B-Instruct model.}
\label{tab:main-longbench}
\end{table*}

\begin{table*}[t]
\centering
\sisetup{
    table-format=2.2,
}
\begin{adjustbox}{width=1.99\columnwidth,center}
\begin{tabular}{lcSSSSSSSSSS}
\toprule
\multicolumn{1}{l}{\textbf{Method}} & \textbf{Avg. bit}  ($\downarrow$) & {\textbf{R.PK}} & {\textbf{R.Num}} & {\textbf{R.KV}} & {\textbf{En.Sum}} & {\textbf{En.QA}} & {\textbf{En.MC}} & {\textbf{En.Dia}} & {\textbf{Code.D}} & {\textbf{Math.F}} & {\textbf{Average} ($\uparrow$)} \\ \midrule
FP16 Baseline & 16 & 100.00  & 99.49 & 55.20 & 26.74 & 14.28 & 66.81 & 20.00 & 22.08 & 33.43 & 48.67 \\ \midrule
KIVI-2 & 3.00 & 100.00 & 97.80 & 0.60 & 25.41 & 13.90 & 66.81 & 22.50 & 23.35 & 33.71 & 42.68 \\
KVQuant-2 & 2.33 & 98.81 & 88.81 & 0.00 & 25.02 & 7.77 & 35.81 & 8.00 & 25.63 & 10.29 & 33.34 \\
VQLLM-2 & 2.00 & 100.00 & 97.96 & 0.00 & 18.27 & 8.09 & 44.54 & 9.50 & 21.83 & 29.71 & 36.66\\
\textbf{{\name}-2} & \textbf{2.00} & 100.00 & 93.39 & 12.20 & 24.14 & 14.57 &  67.25 & 18.00 & 22.08 & 33.14 & \textbf{42.75} \\ \midrule
KIVI-1 & 2.00 &  1.86 & 0.00 & 0.00 & 12.44 & 3.24 & 55.46 & 4.50 & 23.86 & 34.00 & 15.04  \\
KVQuant-1 & 1.33 & 0.00 & 0.00 & 0.00 & 17.83 & 1.36 & 0.44 & 2.00 & 0.25 & 1.14 & 2.56 \\ 
VQLLM-1 & \textbf{1.00} & 82.37 & 14.75 & 0.00 & 10.46 & 2.69 & 25.33 & 1.50 & 21.83 & 7.43 & 18.48 \\
\textbf{{\name}-1} & 1.03 & 99.15 & 62.37 & 0.00 & 19.34 & 11.30 & 65.50 & 18.00 & 22.08 & 33.14 & \textbf{36.76} \\ \bottomrule
\end{tabular}
\end{adjustbox}
\caption{InfiniteBench evaluation for the LLaMA-3.1-8B-Instruct model.}
\label{tab:main-infbench}
\end{table*}

\noindent\textbf{Tasks.} To evaluate the effectiveness of our method for long-context LLMs, we test it alongside the baselines on two long-context benchmarks: LongBench~\citep{bai2023longbench} and InfiniteBench~\citep{zhang2024bench}. Additionally, to assess the model's ability to perform complex reasoning, we evaluate it on GSM8K~\citep{cobbe2021training}. Apart from the task score, we also report the average quantization bit, denoted as \textbf{Avg. bit}, for each method to quantify the actual KV cache size reduction. A lower \textbf{Avg. bit} indicates less storage required for the KV cache, leading to greater memory savings. For baseline methods, we follow the calculations provided in their respective papers to determine the \textbf{Avg. bit}.

\subsection{Long Context Benchmarks Evaluation}

\noindent\textbf{LongBench Evaluation.} LongBench~\citep{bai2023longbench} is a benchmark for evaluating models on long-context tasks like multi-doc QA, summarization, and code completion. Following KIVI~\citep{liu2024kivi}, we assess performance on the same eight tasks across four subgroups and report both individual and average scores. The maximum sequence length is set to 128K.

Experiment results are presented in Table~\ref{tab:main-longbench}. Our 2-bit quantization model achieves lossless accuracy on most tasks, maintaining almost the same average score as FP16 model while outperforming other baselines as well as offering greater memory savings. Compared to KIVI, our method provides 33\% more memory savings while achieving a higher average score. Among baselines with a similar average quantization bit, our approach performs significantly better, with scores 2.63\% higher than KVQuant and 2.53\% higher than VQLLM. For 1-bit quantization, our method substantially outperforms other baselines, achieving a 17.52\% higher average score than VQLLM while maintaining comparable memory savings. This minimizes accuracy degradation compared to the FP16 model.

\noindent\textbf{InfiniteBench Evaluation.} InfiniteBench~\citep{zhang2024bench} evaluates models on ultra-long contexts, mimicking real-world scenarios with near-infinite input lengths. Its tasks include multiple types of retrieval, QA, summarization, and code debugging. In our experiment, the maximum sequence length is set to 128K.

Experiment results are shown in Table~\ref{tab:main-infbench}. Our method continues to excel in ultra-long context scenarios, with even greater superiority in 1-bit quantization. For challenging tasks that require accurate information, such as retrieval (R.PK, R.Num, and R.KV), other methods fail to produce accurate results, whereas ours retains some capacity even at low quantization levels. This is due to our codebook design, which effectively preserves information in the KV cache, as confirmed by our method’s lower quantization error (measured in MSE) shown in Table~\ref{tab:quant-acc} in Appendix~\ref{sec:quant_error_analysis}.

\begin{figure*}[t]
    \centering
    \includegraphics[width=0.97\textwidth]{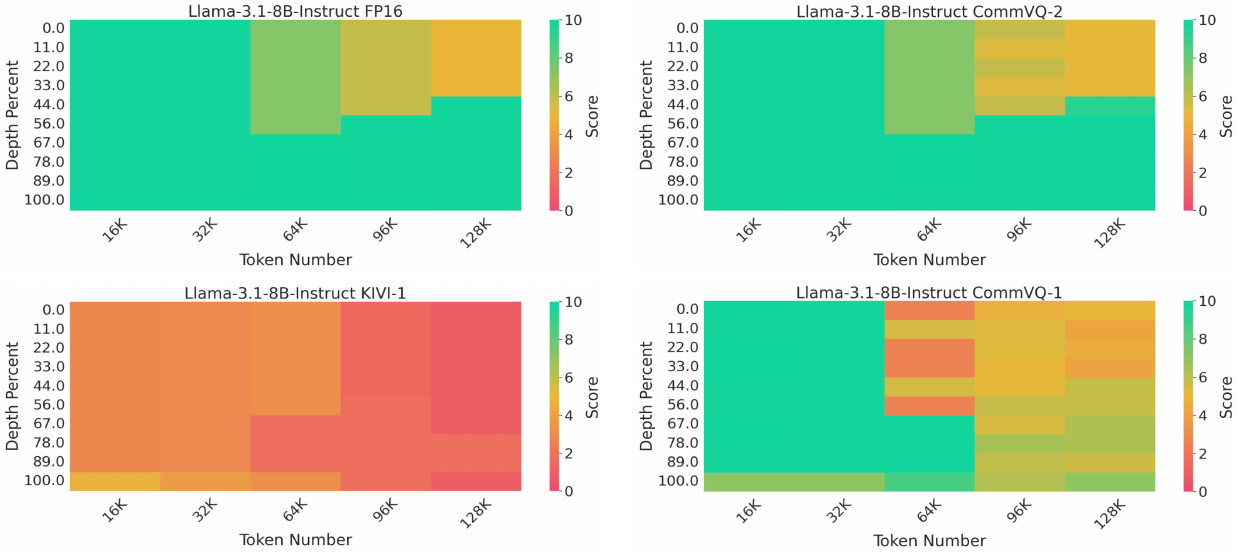}
    \caption{Needle-in-a-Haystack test using the LLaMA-3.1 8B model. We present the test result for FP16 baseline (top-left), KIVI-1 (bottom-left), CommVQ-2 (top-right) and CommVQ-1 (bottom-right). Our method's 2-bit version matches FP16 performance, while the 1-bit CommVQ variant outperforms KIVI, demonstrating strong retrieval fidelity under extreme compression.}
    \label{fig:niah}
\end{figure*}

\noindent\textbf{Needle-in-a-Haystack Evaluation.} We further evaluate our method using LLaMA-3.1 8B model on the Needle-in-a-Haystack (NIAH) benchmark, which specifically targets retrieval capabilities in long-context settings by requiring the model to identify a small piece of information embedded within a large amount of irrelevant text. This test serves as a strong indicator of how well KV cache and attention mechanisms are preserved under quantization.

Experimental results in Figure~\ref{fig:niah} demonstrate that our 2-bit quantization model successfully retains full retrieval capability, matching the performance of the FP16 baseline. This confirms that our approach introduces negligible degradation even under aggressive compression. More notably, our 1-bit CommVQ variant achieves stronger retrieval accuracy than KIVI's 1-bit counterpart, highlighting the effectiveness of our codebook design in preserving critical attention signals. These findings reinforce that CommVQ maintains high fidelity in information-dense retrieval tasks, even under extreme compression ratios.

\begin{table}[t]
\centering
\sisetup{
    table-format=2.2,
}
\begin{adjustbox}{width=0.8\columnwidth,center}
\begin{tabular}{lcS}
\toprule
\multicolumn{1}{l}{\textbf{Method}} & \textbf{Avg. bit} ($\downarrow$) &  {\textbf{GSM8K} ($\uparrow$)} \\ \midrule
FP16 Baseline & 16 & 76.27 \\ \midrule
KIVI-2 & 3.00 & 73.69   \\
VQLLM-2 & 2.00 & 52.69 \\
\textbf{{\name}-2} & \textbf{2.00} & \textbf{76.04}  \\ \midrule
KIVI-1 & 2.00 & 2.20 \\
VQLLM-1 & \textbf{1.00} & 1.67 \\
\textbf{{\name}-1} & 1.03 &  \textbf{66.57} \\ \bottomrule
\end{tabular}
\end{adjustbox}
\caption{GSM8K benchmarks evaluation for the LLaMA-3.1-8B-Instruct model. Exact match accuracy is used as the metric.}
\label{tab:main-general}
\end{table}

\subsection{GSM8K Evaluation}

To demonstrate the effectiveness of our approach on challenging and complex tasks, we conduct experiments on GSM8K~\citep{cobbe2021training}, a rigorous benchmark comprising high-quality, linguistically diverse math problems meticulously crafted by human experts. These problems require intricate multi-step reasoning and involve complex arithmetic operations, making GSM8K an ideal benchmark for evaluating complex reasoning capabilities. As shown in Table~\ref{tab:main-general}, our 2-bit quantization model outperforms other baselines, maintaining significantly higher accuracy, 2.35\% above KIVI and 23.35\% above VQ-LLM, while exhibiting only a minimal accuracy degradation of 0.23\% compared to FP16. Furthermore, while our baselines struggle to generate accurate results under 1-bit quantization, our method continues to demonstrate strong reasoning capabilities even under this extreme compression.

\subsection{Model Ablation}

We also apply our method to two additional long-context LLMs: LLaMA-2-7B (the 32K context length version from Together.ai) and Mistral-7B-v0.3 (which natively supports 32K context length). We evaluate both on LongBench, comparing their average scores to the FP16 baseline and KIVI.

\begin{table}[t]
\centering
\begin{adjustbox}{width=0.95\columnwidth,center}
\begin{tabular}{clccc}
\toprule
\textbf{Model} &  & \textbf{Avg. bit} ($\downarrow$) & \textbf{LongBench} ($\uparrow$) \\ \midrule
\multirow{3}{*}{Llama-2-7B} & FP16 & 16 & 48.43 \\
 & KIVI & 3.00 & 47.14 \\
 & {\name} & \textbf{2.00} & \textbf{47.27} \\ \midrule
\multirow{3}{*}{Mistral-7B} & FP16 & 16 & 53.40 \\
 & KIVI & 3.00 & 52.78 \\
 & {\name} & \textbf{2.00} & \textbf{53.04} \\ \bottomrule
\end{tabular}
\end{adjustbox}
\caption{Performance comparison of full precision (FP16), KIVI, and CommVQ applied to two additional LLMs for model ablation.}
\label{tab:ablation-model}
\end{table}

The experimental results are summarized in Table~\ref{tab:ablation-model}. For both the LLaMA-2 and Mistral models, our method consistently preserves the FP16 baseline's average score while achieving a better compression rate–accuracy trade-off than KIVI. These results highlight the broad applicability and effectiveness of our approach.

\begin{table}[t]
\centering
\begin{tabular}{lccc}
\toprule
Latency (ms) & 8K & 32K & 128K \\ \midrule
Naive Impl. & 2.4 & 9.2 & 36.6 \\
Optimized Impl. & 0.4 & 1.1 & 3.8 \\ \midrule
\textbf{Speedup} & \textbf{6.0} & \textbf{8.4} & \textbf{9.6} \\
\bottomrule
\end{tabular}
\caption{Latency comparison between the naive implementation and the optimized implementation utilizing the commutative codebook. Latency per layer per token is measured for context lengths of 8K, 32K, and 128K, reported in milliseconds (ms).}
\label{tab:speedup}
\end{table}

\subsection{Robustness Analysis Under Domain Shift}

Our codebook and encoder are trained on the FineWeb-Edu~\citep{lozhkov2024fineweb-edu} pre-training dataset. To evaluate their robustness when the testing domain differs from the training domain, we conducted an analysis using the LLaMA-3.1 8B model. Specifically, we compared the perplexity of CommVQ-2 against the FP16 baseline across four distinct datasets: \textbf{FineWeb-Edu}, a general text dataset, also the training set; \textbf{GSM-8K}~\citep{cobbe2021training}, a math dataset; \textbf{Repobench-p}~\citep{bai2023longbench}, a code retrieval and completion dataset; and \textbf{$\mathbf{KV_{\mathrm{Retrieval}}}$ in InfiniteBench}~\citep{zhang2024bench}, a synthetic UUID key-value retrieval dataset. The first dataset represents in-domain evaluation, while the last three represent evaluations with domain shifts, \textit{i.e.}, the codebooks and encoder are trained on general text and tested on math, code, and synthetic UUID data. The results are summarized in Table~\ref{tab:domain_shift}.

\begin{table}[t]
\centering
\resizebox{\linewidth}{!}{
\begin{tabular}{lcccc}
\toprule
\textbf{Method} & \textbf{FineWeb-Edu} & \textbf{GSM-8K} & \textbf{Repobench-p} & \textbf{KV\_Retrieval} \\
\midrule
FP16 & 10.17 & 5.67 & 2.20 & 31.93 \\
CommVQ-2 & 11.54 & 6.14 & 2.78 & 32.72 \\ \midrule
\textbf{PPL Diff} & +1.37 & +0.47 & +0.58 & +0.79 \\
\bottomrule
\end{tabular}
}
\caption{Perplexity (PPL) comparison between FP16 baseline and CommVQ-2 across different domains.}
\label{tab:domain_shift}
\end{table}

We find no significant increase in perplexity (PPL) due to domain shifts when compared to in-domain evaluations. This suggests that our method performs consistently well across domains that differ from the calibration data, including synthetic UUID data, which is unlikely to appear in the calibration set. Overall, we conclude that our method is robust and generalizable under domain shifts.

\subsection{Efficiency Results}

In Table~\ref{tab:speedup}, we present a latency comparison to demonstrate the impact of our method's reduced computation (denoted as the optimized implementation) compared to the naive implementation, which does not utilize the commutative codebook to reduce the computation. For context lengths of 8K, 32K, and 128K, the optimized implementation consistently achieves speedups over the naive implementation, validating the commutative codebook's effectiveness in reducing computational overhead.

\begin{figure}[t]
    \centering
    \begin{minipage}{0.5\textwidth}
        \centering
        \includegraphics[width=\textwidth]{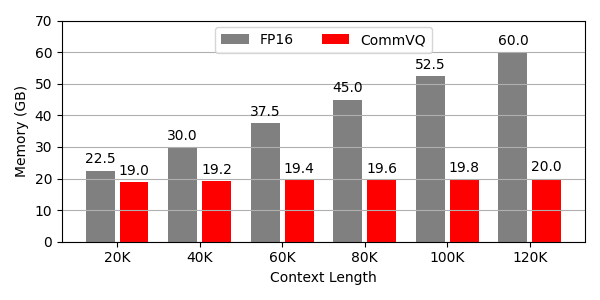} 
        \small{(a) Memory usage vs. context length (batch size = 1).}
        \label{fig:memory1}
    \end{minipage}\vfill
    \begin{minipage}{0.5\textwidth}
        \centering
        \includegraphics[width=\textwidth]{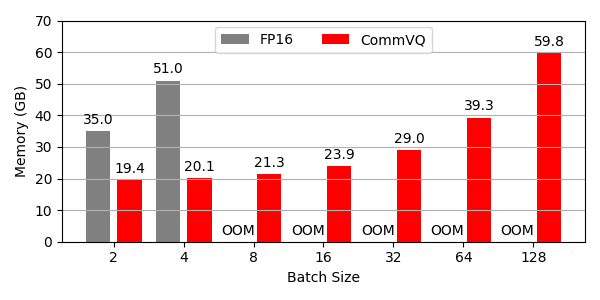} 
        \small{(b) Memory usage vs. batch size (context length = 32K).}
        \label{fig:memory2}
    \end{minipage}
    \caption{\textbf{Per-token decoding memory usage} of {\name} (1-bit) compared to FP16 model. Experiments are conduct on LLaMA-3.1-8B-Instruct model.}
    \label{fig:efficiency}
\end{figure}

We also implement Triton kernels to achieve real memory savings. Figure~\ref{fig:efficiency} highlights the real per-token decoding memory savings using the LLaMA-3.1-8B-Instruct model, measured on an H100-80GB GPU. A 120K context length requires 60GB in FP16, while our method reduces it to 20GB, enabling inference on a single consumer GPU such as RTX 4090. For a 32K context length, FP16 runs into OOM at a batch size of 8, but our method scales up to 128. This improves long-context and large-batch serving, benefiting a lot of applications such as long document QA. We also provide additional analysis of codebook size in Appendix~\ref{sec:codebook_size}, demonstrating that its size is negligible, especially compared to the large KV cache for long contexts.

\section{Conclusion}

We introduce CommVQ, a novel KV cache quantization approach for long-context LLMs. By leveraging vector quantization and a RoPE-commutative codebook, CommVQ significantly reduces KV cache size while maintaining high computational efficiency. Evaluations on long-context benchmarks show that CommVQ outperforms existing KV cache quantization methods, enabling more memory-efficient and scalable long-context LLM inference.

\newpage

\section*{Impact Statement}

This paper presents work whose goal is to advance the field of 
Machine Learning. There are many potential societal consequences 
of our work, none which we feel must be specifically highlighted here.

\nocite{langley00}

\bibliography{main}
\bibliographystyle{icml2025}

\newpage
\appendix
\onecolumn

\section{Appendix}

\subsection{Explanation of Encoding and Decoding Using Commutative Codebooks}
\label{sec:more_detail}

In this section, we provide an additional explanation of how the 2-dimensional sub-vector \( k_i^j \) is represented using the quantized vector \( s_i^j \) and the codebook \( \mathcal{C}_K^{j} \), as defined in Eqn.~\ref{eqn:commute_decode}.

We first formulate the quantization process as a clustering problem, where the cluster centers are defined as follows:
\begin{equation}
    c_{a,b} = \begin{bmatrix}1 \\0\end{bmatrix}\mathcal{C}_K^j[a] + \begin{bmatrix}0 \\1\end{bmatrix}\mathcal{C}_K^j[b]
    \label{eqn:def_center}
\end{equation}
Here, \( \mathcal{C}_K^j[a] \) and \( \mathcal{C}_K^j[b] \) represent the \( a^{th} \) and \( b^{th} \) \( 2 \times 2 \) sub-codebooks in \( \mathcal{C}_K^j \). Consequently, the codebook \( \mathcal{C}_K^j \) forms a total of \( N_{c'}^2 \) clustering centers. We can then quantize $k_i^j$ into its nearest clustering center, and use $s=\{a,b\}$ as the quantized representation to represent $k_i^j$. When decoding, we use the assigned clustering center to approximate $k_i^j$, so
\begin{equation}
\begin{aligned}
    \label{eqn:appendix_comm_decoding}
    \hat k_i^j  &= c_{a,b} \\
                &= \begin{bmatrix}1 \\0\end{bmatrix}C_K^j[a] + \begin{bmatrix}0 \\1\end{bmatrix} C_K^j[b]
\end{aligned}
\end{equation}

And Eqn.~\ref{eqn:appendix_comm_decoding} is doing exactly the same thing as Eqn.~\ref{eqn:commute_decode}.

\subsection{EM Algorithm Implementation Details}
\label{sec:em_algo_impl_details}

As mentioned in Sec~\ref{sec:rope_aware_key}, we apply a EM-like algorithm to learn the codebook. In this section we provide the algorithm detail. We use a subset of FineWeb-Edu~\citep{lozhkov2024fineweb-edu} as the calibration set $K$ and optimize the codebook over this calibration set. The \textbf{E Step} is straightforward as it simply assign each vector $k$ in $K$ to its nearest clustering center. For \textbf{M Step}, we derive a closed form formula to update $\mathcal{C}_K^j$ given the current assignment $S$. Recall the definition of $\mathcal{C}_K^j$:
\begin{align}
    \mathcal{C}_K^j &= \{C_K^{j0}, C_K^{j1}, \ldots, C_K^{j(N_{c'}-1)}\} \\
    C_K^{jl} &= \begin{pmatrix}
    x_l & y_l \\
    -y_l & x_l \\
\end{pmatrix}
\end{align}

We first define some useful terms:
\begin{equation}
\begin{aligned}
\label{eqn:useful_terms}
\boldsymbol{\phi} &= [x_i, y_i]_{i=0}^{N_{c'}-1} \\
\boldsymbol{m} &= [m_{ij}]_{i,j=0}^{N_{c'}-1, N_{c'}-1} \\
\boldsymbol{S} &= \operatorname{diag}(N_{ij}, N_{ij})_{i,j=0}^{N_{c'}-1,N_{c'}-1}
\end{aligned}
\end{equation}
where $\boldsymbol{\phi}$ is a $2N_{c'}$-dimensional vector that represent the codebook $\mathcal{C}_K^j$ and we are going to optimize it. $\boldsymbol{m}$ is a $N_{c'}^2$-dimensional vector that represents the mean for the data points assigned to each cluster center. $\boldsymbol{S}$ is a $2N_{c'}\times 2N_{c'}$ matrix that has the total number of data points assigned to each cluster center in its diagonal.

Next, we define an auxiliary constant matrix $T\in\{-1,0,1\}^{(2{N_{c'}}^2)\times (2N_{c'})}$, where its entries are given by:

\begin{align}
T_{2(xN_{c'} + y), 2x} = 1, \quad T_{2(xN_{c'} + y), 2y+1} = -1 \\
T_{2(xN_{c'} + B) + 1, 2y} = 1, \quad T_{2(xN_{c'} + y) + 1, 2x+1} = 1 \\
\end{align}

for all \( x, y \in \{0, 1, \dots, N_{c'}-1\} \), and all other elements of $T$ are zero.

We can then rewrite Eqn.~\ref{eqn:commute_decode} into the matrix form:
\begin{equation}
    \min_{\boldsymbol{\phi}} (\boldsymbol{T}\boldsymbol{\phi}-\boldsymbol{m})^T\boldsymbol{S}(\boldsymbol{T}\boldsymbol{\phi}-\boldsymbol{m})
\end{equation}

The closed form solution is given by
\begin{equation}
    \boldsymbol{\phi}^* = (\boldsymbol{T}^T\boldsymbol{S}\boldsymbol{T})^{-1}\boldsymbol{T}^T\boldsymbol{S}\boldsymbol{m}
\end{equation}
We iteratively update $\boldsymbol{\phi}$ until it converges. During our experiments, we found that the learning process was not stable and it would frequently fail to optimize since the number of the clustering centers is large in our case (\textit{e.g.}, there are $4096$ clustering centers for$N_{c'}=64$) and the clustering centers needs to satisfy Eqn.~\ref{eqn:def_center}. In order to stabilize the optimization process, we employ two techniques. The first technique is \textbf{soft clustering center assignment}, where instead of hard-assigning a data point to its nearest clustering center, we distribute it among all clustering centers with different weights. The weight assigned to each center depends on its proximity to that data point, with the closest center receiving the highest weight. To be specific, let $N$ be the number of data points in the calibration set $K$, and $N_{cc}$ be the number of clustering centers, weight matrix $W\in\mathbb{R}^{N \times N_{cc}}$ is calculated as
\begin{equation}
\label{eqn:softmax_weight}
W_{ij} = \frac{e^{-D_{ij}}}{\sum_k e^{-D_{ik}}}
\end{equation}
where $D\in\mathbb{R}^{N \times N_{cc}}$ is the L2 distance matrix measuring the L2 distance between each data point and each clustering center. We then use $W$ instead of the hard assignment to calculate $\boldsymbol{m}$ and $\boldsymbol{S}$ in Eqn.~\ref{eqn:useful_terms}.

Furthermore, we empirically observe that in the early iterations, the distribution of \( W \) needs to be smoother to prevent dead clustering centers. In contrast, in later iterations, a sharper distribution of \( W \) helps achieve better convergence. Therefore, we introduce \textbf{temperature annealing} in Eqn.~\ref{eqn:W} to regulate the distribution:
\begin{equation}
\label{eqn:W}
W_{ij} = \frac{e^{-\frac{D_{ij}}{T}}}{\sum_k e^{-\frac{D_{ik}}{T}}}
\end{equation}
where \( T \) is the temperature parameter, which exponentially decays over the iterations.

\subsection{Codebook Size Analysis}
\label{sec:codebook_size}

Our method uses vector quantization with a codebook to compress the KV cache, which requires additional GPU memory to store the codebook. The codebook configuration is shown in Table~\ref{tab:codebook_conf} and is stored in FP16. The total codebook size is calculated as:

\begin{align}
    \text{Value Codebook Size (MB)} &= N_c\times d \times 2 \\
    \text{Key Codebook Size (MB)} &= 2\times 2\times N_{c'}\times R \times \frac{d}{2} \times 2
\end{align}

where $d$ is the hidden size of the LLM. For LLaMA-3.1-8B-Instruct model, $d=1024$. $N_c$ is the number of rows in the value codebook. For the key codebook, $N_{c'}$ means the number of quantization level, $R$ means the number of the residual quantization. 

\begin{table}[ht]
    \centering
    \begin{minipage}{0.9\columnwidth}
        \centering
        \begin{tabular}{lcc}
        \toprule
        \multicolumn{1}{c}{} & \multicolumn{2}{c}{\textbf{Avg. Bit}} \\
        \multicolumn{1}{c}{} & 1 bit & 2 bit \\ \midrule
        Value Codebook & 2.00 MB & 4.00 MB \\
        Key Codebook   & 2.75 MB & 5.25 MB \\ \bottomrule
        \end{tabular}
        \caption{\textbf{Analysis on Codebook Size.} We calculate the codebook size based on LLaMA-3.1-8B-Instruct model.}
        \label{tab:codebook_size}
    \end{minipage}
    \vfill
    \vspace{7mm}
    \begin{minipage}{0.9\columnwidth}
        \centering
        \begin{tabular}{lcc}
        \toprule
        \multicolumn{1}{c}{} & 1 bit & 2 bit \\ \midrule
        $N_c$ & 1024 & 2048 \\
        R & 11 & 21 \\
        $N_{c'}$ & 64 & 64 \\ \bottomrule
        \end{tabular}
        \caption{\textbf{Codebook Configuration.} $N_c$ is used for value codebook, and $R,N_{c'}$ is used for key codebook.}
        \label{tab:codebook_conf}
    \end{minipage}
\end{table}

We analyze the codebook size for both 2-bit and 1-bit quantization for LLaMA-3.1-8B-Instruct model, as shown in Table~\ref{tab:codebook_size}. For comparison, a KV cache with a 128K context length requires 256 MB each for values and keys. Notably, the codebook size remains constant regardless of the number of tokens, making its memory overhead negligible for long-context LLM inference.

\subsection{Ablation Study on Commutative Codebook Configuration}
\label{sec:ablate_codebook}

We utilize a specially designed commutative codebook to quantize key cache as described in Sec.~\ref{sec:rope_aware_key}. There are three hyper-parameters that control the level of compression: $N_{c'}$, the number of quantization levels, $R$, the number of the residual quantization, and $g$, the number of sub-vector in a group that share the same quantization vector $s$. Revisit that the formula to compute the average quantization bit is:
\begin{equation}
    \textbf{Avg. bit} = \frac{R\log_2(N_{c'})}{g}
\end{equation}

All experiments in this section are conducted using LLaMA-3.1-8B-Instruct model and the quantization error, measured in MSE, is calculated using the key cache for the first layer of the LLaMA model on a subset of the FineWeb-Edu~\citep{lozhkov2024fineweb-edu} dataset.

In our first ablation study, we examine the effect of $g$ when we make $R=1$ and keep the average quantization bit to be the same, as shown in Table~\ref{tab:vary_g}. We also vary $R$ for different $g$ to keep the average quantization bit to be 1, as shown in Table~\ref{tab:vary_r}.

\begin{table}[h]
    \centering
    \begin{minipage}{0.4\columnwidth}
        \centering
        \begin{tabular}{cccc}
        \toprule
        g & $N_{c'}$ & R & \textbf{MSE} \\ \midrule
        8 & 2 & 1 & 0.2699 \\
        16 & 4 & 1 & 0.2011 \\
        32 & 16 & 1 & 0.1265 \\
        64 & 64 & 1 & \textbf{0.0906} \\ \bottomrule
        \end{tabular}
        \caption{Ablation study on $g$ with $R=1$, while maintaining a consistent \textbf{Avg. bit}.}
        \label{tab:vary_g}
    \end{minipage}
    \hfill
    \begin{minipage}{0.4\columnwidth}
        \centering
        \begin{tabular}{ccccc}
        \toprule
        g & $N_{c'}$ & R & \textbf{Avg. bit} & \textbf{MSE} \\ \midrule
        8 & 2 & 8 & 1 bit & 0.0798 \\
        16 & 4 & 8 & 1 bit & 0.0790 \\
        32 & 16 & 8 & 1 bit & 0.0254 \\
        64 & 64 & 11 & 1 bit & \textbf{0.0095} \\ \bottomrule
        \end{tabular}
        \caption{Ablation study on $g$ and $R$ while keeping the \textbf{Avg. bit} to be 1.}
        \label{tab:vary_r}
    \end{minipage}
\end{table}

From both Table~\ref{tab:vary_g} and~\ref{tab:vary_r} we can conclude that when keeping the average quantization bit the same, a larger $g$ will result in a lower quantization error, though in the cost of an increased number of $N_{c'}$ which will induce an increase in the computation complexity as stated in Eqn.~\ref{eqn:final_comp}. Comparing the same row in Table~\ref{tab:vary_g} and~\ref{tab:vary_r} that has the same $g$ and $N_{c'}$, we can further conclude that a larger $R$ will bring lower quantization error when keeping $g$ and $N_{c'}$ unchanged.

\begin{figure}[h]
    \centering
    \includegraphics[width=0.4\columnwidth]{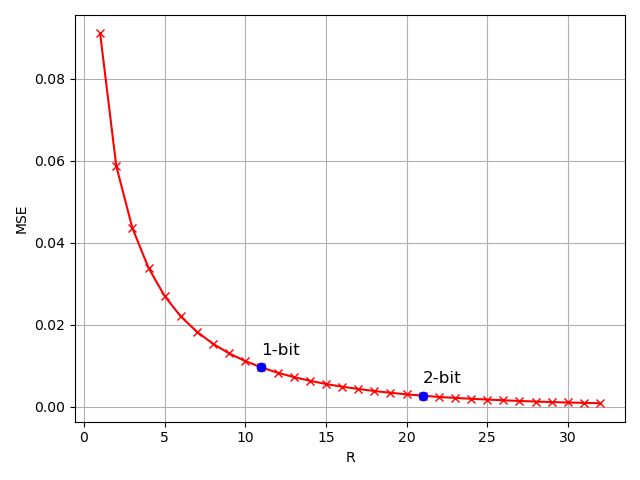}
    \caption{Ablation study on $R$ while keeping $g=64$ and $N_{c'}=64$.}
    \label{fig:error_vs_residual}
\end{figure}

Next, we keep $g=64$ and $N_{c'}=64$, and vary $R$ to get the quantization error for different compression rate. The result is shown in Figure~\ref{fig:error_vs_residual}. The $R$ value for 1-bit quantization and 2-bit quantization is labeled in the figure by a blue dot. As we can see, increasing $R$ will continuously decrease the quantization error, in the cost of a higher average quantization bit.

In conclusion, a larger $g$ and larger $R$ will lead to better quantization accuracy, but also lead to higher computation and lower compression rate. As a result, to achieve a good trade-off between quantization accuracy, computation cost and compression rate, we set $g=64$, $N_{c'}=64$ for all our main experiments, and $R=11$ for 1-bit quantization and $R=21$ for 2-bit quantization respectively.

\subsection{Quantization Error Comparison}
\label{sec:quant_error_analysis}

We conduct experiments to compare the quantization error of our method, measure in MSE between the original KV cache and the decoded KV cache, with our baseline method. To be specific, we use the value cache from the first layer of LLaMA-3.1-8B-Instruct model for MSE calculation. Asymmetric quantization serves as the baseline for comparison. From Table~\ref{tab:quant-acc}, we observe that our additive quantization-based method outperforms asymmetric quantization, particularly when the quantization bit is low, such as in the case of 1-bit quantization.

\begin{table}[h]
\centering
\begin{tabular}{lcc}
\toprule
\multicolumn{1}{l}{\textbf{Method}} & \textbf{Avg. bit} & \textbf{MSE} \\ \midrule
Asymmetric quant. & \multirow{2}{*}{2 bit} & 0.00030 \\
\textbf{{\name}} &  & \textbf{0.00014} \\ \midrule
Asymmetric quant. & \multirow{2}{*}{1 bit} & 0.00380 \\
\textbf{{\name}} &  & \textbf{0.00027} \\ \bottomrule
\end{tabular}
\caption{Comparison of MSE between {\name} and asymmetric quantization used in KIVI~\citep{liu2024kivi}, calculated using the cached value matrix from the first layer of LLaMA-3.1-8B-Instruct. The MSE is evaluated on a small subset of the FineWeb-Edu dataset~\citep{lozhkov2024fineweb-edu}.}
\label{tab:quant-acc}
\end{table}


\end{document}